\documentclass{article}
\usepackage[utf8]{inputenc}
\usepackage{graphicx}
\usepackage{fullpage}
\usepackage{authblk}
\usepackage{amsmath}
\usepackage{hyperref}
\usepackage{enumitem}
\usepackage{subcaption}
\setlist{nolistsep}

\title{DeepMovie: Using Optical Flow and Deep Neural Networks to Stylize Movies}
\author[1]{Alexander G. Anderson}
\author[1]{Cory P. Berg}
\author[1]{Daniel P. Mossing}
\author[1]{Bruno A. Olshausen}
\affil[1]{Redwood Center for Theoretical Neuroscience, University of California, Berkeley}

\begin{document}

\maketitle

\begin{abstract}
A recent paper by Gatys et al. \cite{gatys2015neural} describes a method for rendering an image in the style of another image. 
First, they use convolutional neural network features to build a statistical model for the style of an image. 
Then they create a new image with the content of one image but the style statistics of another image.
Here, we extend this method to render a movie in a given artistic style. The naive solution that independently renders each frame produces poor results because the features of the style move substantially from one frame to the next. The other naive method that initializes the optimization for the next frame using the rendered version of the previous frame also produces poor results because the features of the texture stay fixed relative to the frame of the movie instead of moving with objects in the scene. The main contribution of this paper is to use optical flow to initialize the style transfer optimization so that the texture features move with the objects in the video. Finally, we suggest a method to incorporate optical flow explicitly into the cost function.  
\end{abstract}

\section{Introduction and Related Research}

Automatic image stylization is a popular feature in image editors like Adobe Photoshop and social networking services like Instagram. However, many of these styles are limited in their complexity. Recent work has shown that we can use statistical correlations of neural network features to describe complex styles. Furthermore, these models can be then used to create new images with the same statistics. But a question remains: how do we apply these complex styles to video? This task is additionally challenging because the textures need to move smoothly with the objects in the scene to obtain an pleasing video. In this section, we review past work that motivated our approach of using optical flow in combination with neural network features to stylize movies. 

In a seminal paper on texture generation, Heeger and Bergen built a statistical model of textures using an image pyramid and then developed a method to render new images with those same statistics \cite{heeger1995pyramid}. While this approach produces good results, the method is limited due to the use of hand-designed features. For alternative methods of texture synthesis, we direct the reader to a pair of papers on texturizing images and a more recent review \cite{efros1999texture, efros2001image, wei2009state}.

Recent research on artificial neural networks has shown that the features produced by task-driven learning are better than hand-designed features. In particular, convolutional neural networks trained on image classification (e.g. AlexNet trained on IMAGENET \cite{krizhevsky2012imagenet}, \cite{russakovsky2015imagenet}) extract image features that are useful for many tasks such as fine grained image recognition, scene recognition, and image retrieval \cite{razavian2014cnn}. In order to understand the information captured by neural network features, a number of papers have used backpropagation to find an image that has a given set of neural network features \cite{mahendran2015understanding, Cimpoi2015}. 

These advances in building useful image features and learning how to manipulate them has led to an explosion of research into generating artistic images using deep neural networks. Two recent notable examples are Google's post on inceptionism \cite{mordvintsev2015inceptionism} and the pair of papers from
Gatys et al. \cite{Gatys2015texture, gatys2015neural}.  
Since then, there have been a number of papers that have sought to explore various aspects of the idea and to speed up the rendering (e.g. using an adversarial network approach) \cite{Johnson2016, ulyanov2016texture, li2016combining, lin2015visualizing, lu2015learning, yin2016content, denton2015deep}. 

In order to apply these methods to movies, the textures that these methods overlay onto an image must move with the objects in the scene. The problem of detecting the motion of objects in a scene is well-studied in the field of computer vision and is often referred to as computing the optical flow. Thus we can use the optical flow fields in order to move the textures with the objects as they move in the movie. 

This paper is organized as follows. First, we review previous research, including the formulation of the Gatys et al. style transfer paper. Next, we extend this method using optical flow to encourage temporal continuity. Then we present our results on rendering a naturalistic movie and assess the success of our method as well as techniques for improving it. Finally, we end with some discussion and conclusions. 

\section{Theoretical Formulation}
\label{theory}

First, we review the approach taken by Gatys et al. \cite{gatys2015neural}. While past work used hand-designed features to build statistical models of textures, Gatys et al. had the idea to use neural network features instead. To be more precise, suppose that we have a convolutional neural network such as VGG-19 \cite{Simonyan14c}.
Denote the input image to the network as $x$ (with the individual pixels of the image denoted as $x_{u,v}$) and the activations of the $l$th convolutional layer for feature map $j$ and pixel $i$ as $F_{i,j}^l(x)$ (suppose that there are $J^l$ feature maps and $I^l$ pixels in that layer). Then define the Gram matrix
\begin{equation}
G_{j,j'}^l(x) = \frac{1}{I^l}\sum_i F_{i,j}^l(x) F_{i,j'}^l(x).
\end{equation}
We can think of this matrix as computing the second order statistics between different neural network features because we are averaging over the spatial dimensions of the feature maps of a convolutional network. Part of the intuition of this approach is that complex, higher-order, statistics in the image space can be captured using second order correlations in the neural network feature space. The style transfer objective function takes in two images $x^s$ and $x^c$, the style and content images, respectively and renders an image, $x^r$:
\begin{align}
x^r &= \text{argmin}_x E(x) \\ 
E(x) &= \mathcal{L}_{content}(x, x^c) + \mathcal{L}_{style}(x, x^s) + \mathcal{L}_{tv}(x)\\
\mathcal{L}_{content}(x, x^c) &= \sum_l \lambda^l_c\frac{1}{I^lJ^l}\sum_{i,j} (F_{i,j}^l(x) - F_{i,j}^l(x^c))^2 \\
\mathcal{L}_{style}(x, x^s) &= \sum_l \lambda^l_s\frac{1}{(J^l)^2}\sum_{j, j'} (G_{j,j'}^l(x) - G_{j,j'}^l(x^s)) ^ 2 \\
\mathcal{L}_{tv}(x) &= \lambda_{tv} \sum_{u,v} (x_{u+1,v}-x_{u,v})^2 + (x_{u,v+1}-x_{u,v})^2
\end{align}
where $\lambda^l_{c, s, tv}$ are weights that determine the relative strength of the style, content, and total variation costs and $x_{u,v}$ denotes a pixel of the image.

Now we describe several methods for extending this work to movies. Suppose that we have a movie, whose frames are $x_t$ for $t=0,1,\ldots$. Since the Gram matrix cost is spatially invariant, there are many ways to overlay a texture onto the scene. In the language of optimization, there are many local minima and the initialization has a strong impact on the result of a gradient descent minimization procedure. We exploit this property to obtain temporal continuity in our movie. Let $y = ST(x_0, x^c, x^s)$ denote the output of a gradient descent based minimization of the style transfer objective function that uses $x_0$ as the initialization point for the optimization, $x^c$ as the content image, and $x^s$ as the style image. 

\begin{figure}
	\begin{center}
		\includegraphics[scale = 0.5]{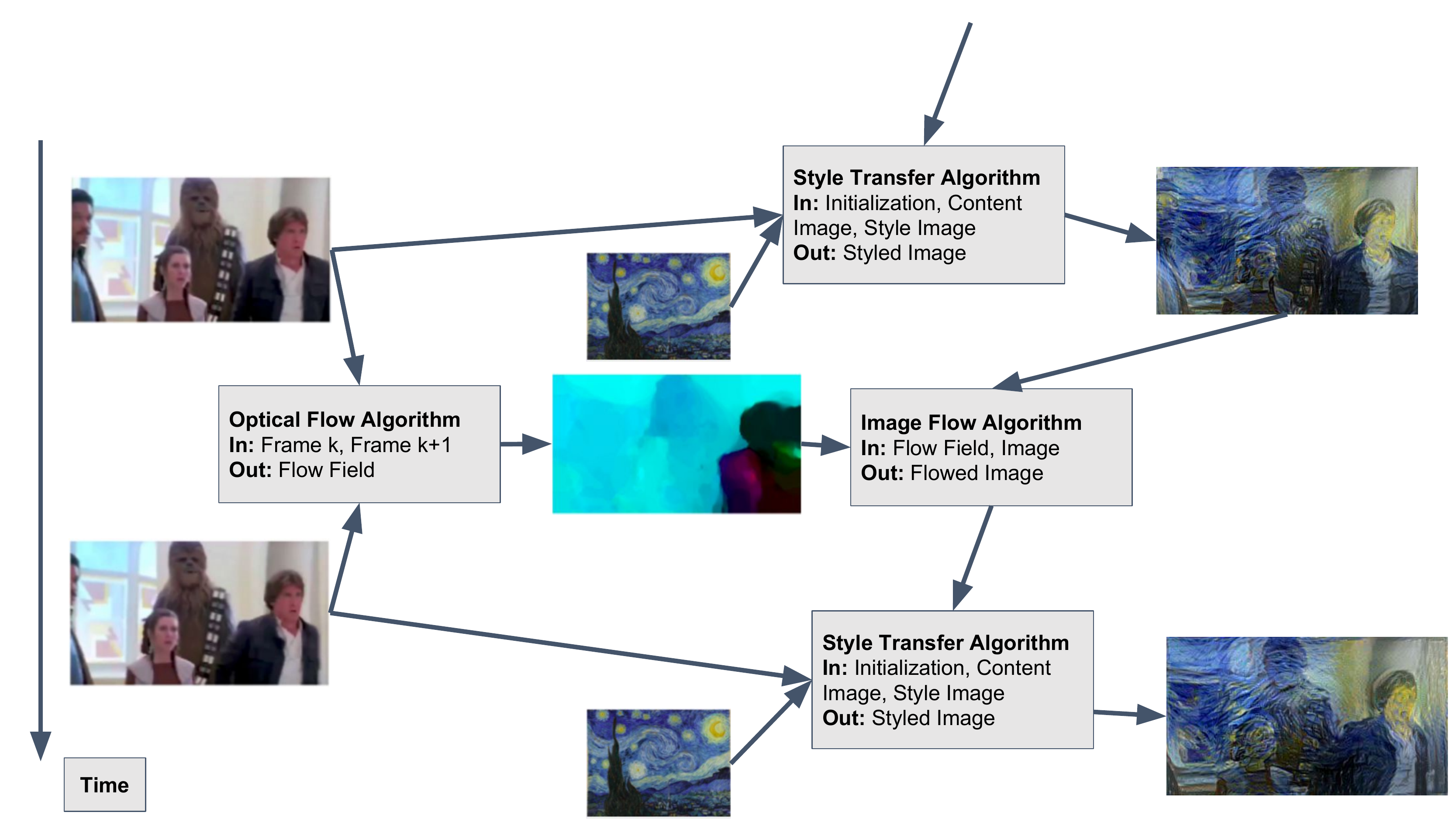}
	\end{center}
	\caption{Overview of Our Approach: We begin by applying the style transfer algorithm to the first frame of the movie using the content image as the initialization. Next, we calculate the optical flow field that takes the first frame of the movie to the second frame. We apply this flow-field to the rendered version of the first frame and use that as the initialization for the style transfer optimization for the next frame. Note, for instance, that a blue pixel in the flow field image means that the underlying object in the video at that pixel moved to the left from frame one to frame two. Intuitively, in order to apply the flow field to the styled image, you move the parts of the image that have a blue pixel in the flow field to the left. }
	\label{fig:approach}
\end{figure}

Alternative methods for applying art style to movies that we implemented (the first two methods are baselines and the third method is the optical flow based method that works better than the baseline methods): 
\begin{itemize}
\item Independent Initialization: $y_t = ST(x_t, x_t, x^s)$. 
\item Rendered Previous Frame Initialization: $y_t = ST(y_{t-1}, x_t, x^s)$.
\item Optical Flow Initialization: $y_t = ST(y_t', x_t, x^s)$, $y_t' = FLOW(y_{t-1}, F_{t-1,t})$ where $F_{t-1,t}$ denotes the optical flow field taking $x_{t-1}$ to $x_t$ and $FLOW$ applies an optical flow field to an image. This approach is summarized in Fig. \ref{fig:approach}. 
\end{itemize}
Additional methods for applying art style to movies that may improve results, but that we have not implemented yet:
\begin{itemize}
\item Optical Flow Initialization and Loss: In addition to doing the above, we can add a term $l(x,y_t')$ that penalizes the distance between the image that we want to render and the optical flowed version of the previous frame of the movie. For $l$, we can either use a robust loss or a squared error loss following the optical flow literature (eg. \cite{sun2010secrets}). 
\item Optical Flow with Back-Tracking: Instead of sequentially optimizing each frame, one can imagine creating an objective function that jointly optimizes over all frames. For instance, we can add the following term:
\begin{equation}
\mathcal{L}_{of, t}(x_t) = \sum_{t'\in N(t)} l(x_t, FLOW(y_{t'}, F_{t',t}))
\end{equation}
where $N(t) = \{t-1, t+1\}$. The full objective function can be descended by pre-computing the optical flow fields and then descending with respect to each image sequentially and then doing additional passes to clean up the images.
\end{itemize}

\section{Results}

In order to show the generality of our technique, we test our approach on a short clip from Star Wars: Episode V - The Empire Strikes Back. This clip has a number of properties that are interesting and challenging for our algorithm: 

\begin{enumerate}
\item There are parts of the video where there is both large motion of the camera and large motion of the objects in the scene. 
\item Surfaces are revealed and hidden as time flows forward. 
\item There are special effects such as rapid flashes of light that create large variations in the overall brightness of the scene, temporarily occlude objects, and break the brightness constancy assumption. 
\end{enumerate}

\begin{figure}
    \centering
    \includegraphics[scale=0.8]{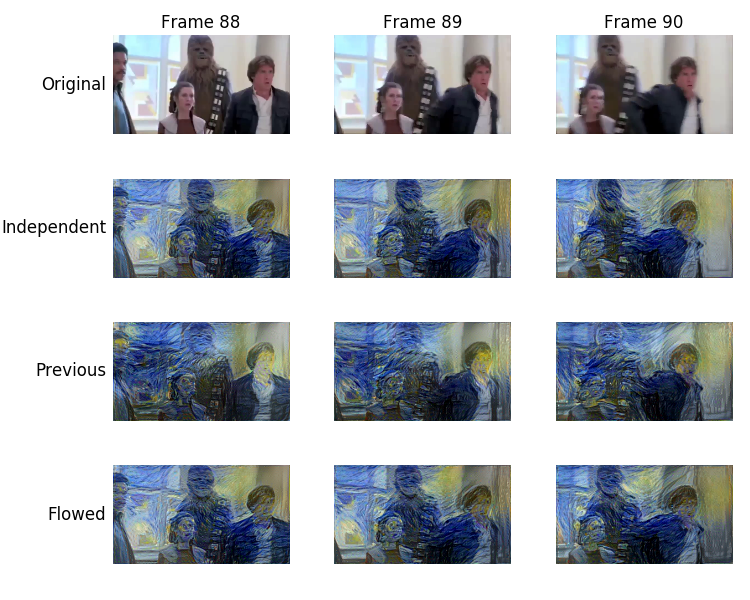}
    \caption{Initialization using optical flow improves results relative to baseline methods during camera panning: Each row shows several adjacent frames of the movie rendered according to a different method. 
    Row One: Frames from the original movie at 10 Hz. 
    Row Two: Independent Initialization - strokes in adjacent frames show little or no apparent relationship.
    Row Three: Previous Rendered Frame Initialization - brush strokes stay fixed relative to the frame of the movie and the characters appearing to melt into the background as the camera pans. 
    Row Four: Flowed Previous Rendered Frame Initialization - strokes appear to move smoothly with the characters.}
    \label{fig:alt}
\end{figure}

\begin{figure}
	\begin{center}
		\includegraphics[scale = 0.5]{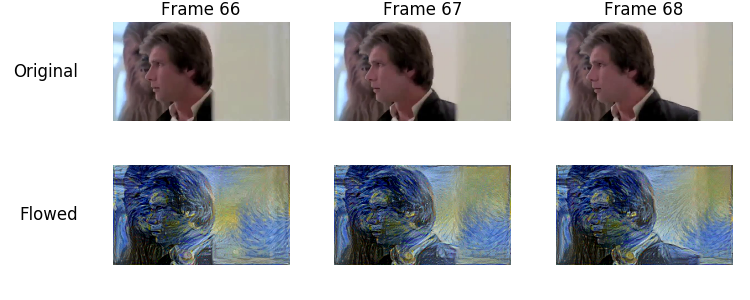}
	    \includegraphics[scale = 0.5]{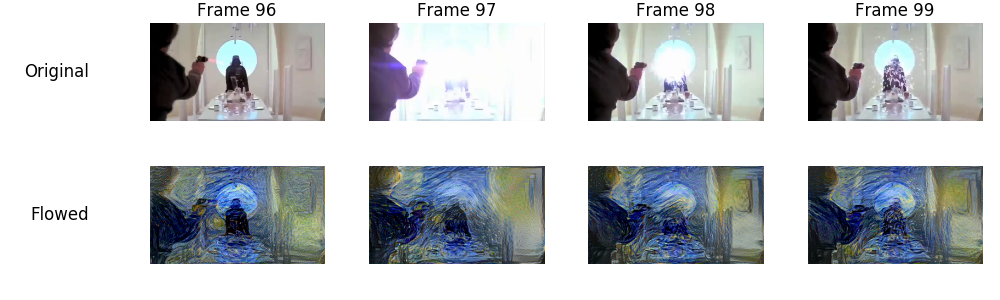}
	\end{center}
	\caption{Optical Flow Method During Occlusions and Special Effects: 
	Top: The content cost appropriately stylizes a surface that is gradually revealed due to a moving occluder.
	Bottom: The resulting rendered video does not capture the special effect and gets warped as a result. }
	\label{fig:flow_results}
\end{figure}

A video summary of our results can be found at \url{https://youtu.be/nOrXeFBkP04}. In Fig. \ref{fig:alt}, we show that our algorithm handles large camera and image motions well in comparison to two baselines: independent rendering of each frame, and initialization using the previous rendered frame. In the case of independent initialization, strokes in adjacent frames show little or no apparent relationship. The previous frame initialization method generates a video where characters appear to melt into the background as the camera pans. In the case of the flowed previous frame initialization, the strokes appear to move smoothly with the characters. In Fig. \ref{fig:flow_results}, we show that occluded surfaces are rendered properly. Also we see that special effects (here, the screen lighting up due to a laser blast) are lost to some extent, but the results still look reasonable. 

\section{Discussion and Conclusions}

Compared to many machine learning tasks that have clear benchmarks, computer generated art is a subjective task. On one hand, a quantitative (and arguably objective) measure of a stylized video is the sum of the squares of the difference between the flowed version of one frame and the next frame. Relatedly, we hypothesize that we could obtain more temporally consistent results by adding an explicit term in the cost function to regularize the style transfer.  On the other hand, some variability from frame to frame contributes to the artistic effect. For instance, a video directed by Boris Seewald makes good use of a combination of temporal continuity in content but discontinuity in style \url{http://www.thisiscolossal.com/2016/03/rotoscoped-music-video-boris-seewald-disco/}. Continued work in this field will benefit from input from artists as to what are desirable properties of stylized movies. 

Along the lines of developing artistically pleasing results, one of the major limitations of this approach is that all of the different style attributes are wrapped up in a model of style that has many parameters, making it difficult to manipulate. For instance, simple properties such as the overall light-level of an image are entangled in the thousands of parameters in the gram-matrix texture model. We experimented with one approach of addressing this issue. In the supplementary figures, we use histogram matching to rescale the style image and then compute a new gram matrix (see Fig. \ref{fig:hist_transfer}). From a practical (eg. for artists) and theoretical perspective, it would be beneficial to figure out how to disentangle simple image properties from the more complex properties (such as brush strokes) that the neural style features capture. 

Another limitation is that in order to make these algorithms commercially useful, the rendering algorithm must be optimized, possibly adapting the approaches suggested in \cite{li2016combining} or \cite{Johnson2016}. The current approach requires a relatively costly gradient descent procedure to render each frame. 

In conclusion, our work demonstrates that we can improve the temporal coherence of an artistically styled movie by making use of optical flow. While statistics based methods of rendering textures are promising, much work needs to be done in order to tease apart simple image properties to make these rendering algorithms useful for artists.

\section{Statement on Contribution}
This project benefited from the collaboration of a number of people. AGA did the literature review, came up with the idea to stylize movies, and implemented the majority of the code (based on rewriting \url{https://github.com/anishathalye/neural-style}). BAO originally suggested using optical flow to make the rendered image more temporally coherent. CPB helped in the debugging of the main algorithm, and ran the controls. DPM used histogram matching to better handle the special effects and contributed to the write-up. We thank Brian Cheung and Dylan Paiton from the Redwood Center for Theoretical Neuroscience and Richard Zhang and Alyosha Efros from the Berkeley Computer Vision Group for useful feedback. 

\bibliographystyle{unsrt}
\bibliography{finalproj}

\section{Supplementary Methods}

\begin{enumerate}

\item The core style transfer algorithm was implemented using TensorFlow \cite{tensorflow2015-whitepaper}. We used a pretrained version of VGG-19 \cite{Simonyan14c} available from \url{http://www.vlfeat.org/matconvnet/models/imagenet-vgg-verydeep-19.mat}.

\item Experimentally we found that quasi-second order methods perform better than vanilla gradient descent. In particular, we use Adam with about 1000 iterations to optimize the objective function for each frame of the movie \cite{kingma2014adam}. Others papers use L-BFGS \cite{liu1989limited} (eg. \cite{gatys2015neural}).  
\item For all experiments, we use $\text{conv4\_2}$ for the content layer and $\text{conv1\_1}$, $\text{conv2\_1}$, $\text{conv3\_1}$, $\text{conv4\_1}$, $\text{conv5\_1}$ for the style layers along the lines of \cite{gatys2015neural}. However, we use different weight values. Roughly speaking, we set the weights so that each term in the cost function contributed at least five percent to the total cost. During optimization, we found that the content cost would still dominate. 
\item In order to generate optical flow fields, we use a pre-compiled binary from T. Brox and J. Malik \cite{brox2011ldof}. We note that these dense flow fields are much better than those produced by the built-in optical flow methods in OpenCV. Our results could be improved even more using a better optical flow algorithm.
\item We use the content image to initialize the optimization in the case of the first frame of the video or in the presence of manually detected scene cuts.  
\item The movies are rendered at 10 frames per second. 
\end{enumerate}

\section{Supplementary Figures}

\begin{figure}
	\begin{center}
		\includegraphics[scale = 0.25]{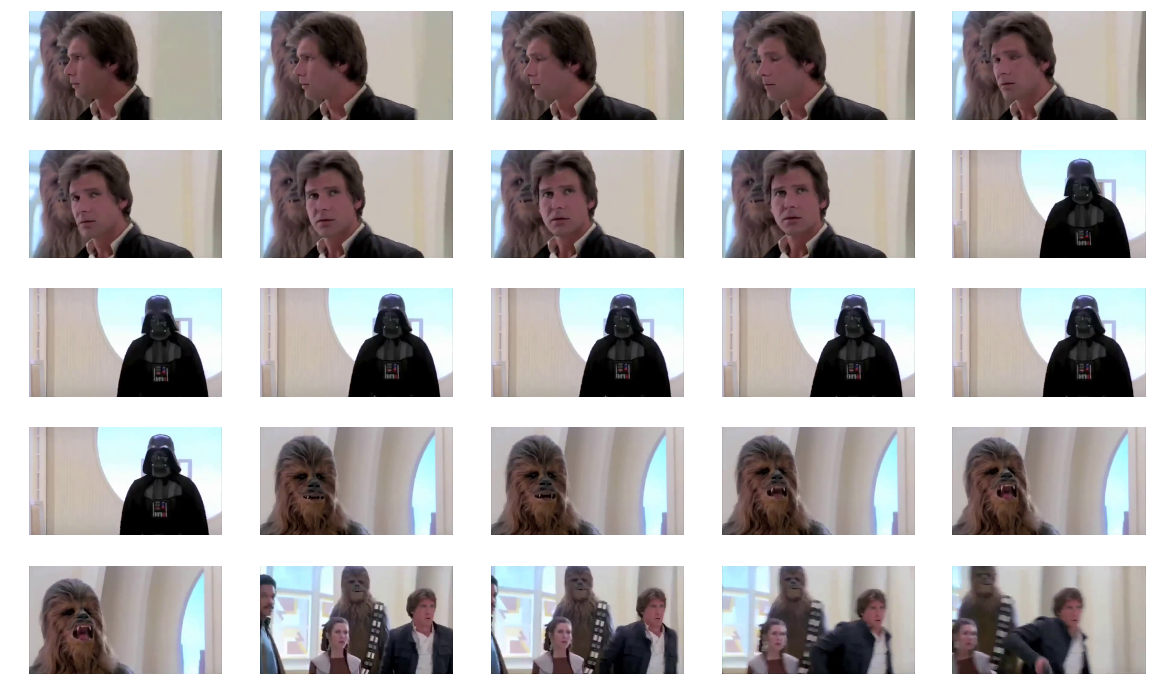}
		\includegraphics[scale = 0.25]{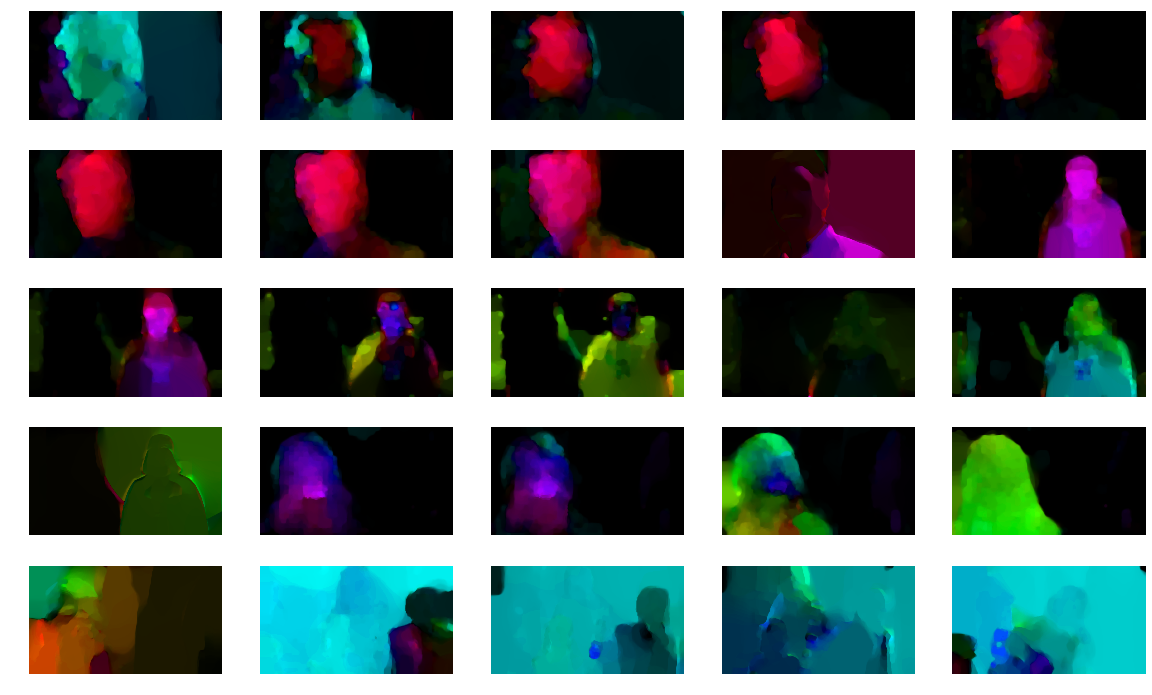}
		\includegraphics[scale = 0.25]{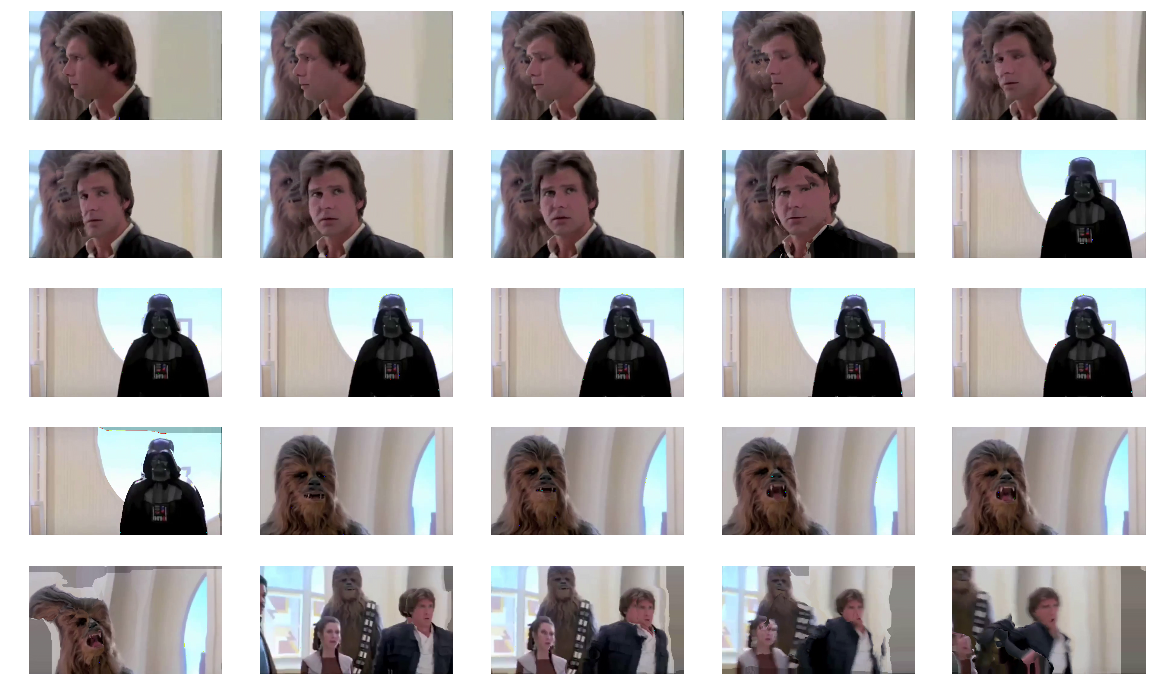}
	\end{center}
	\caption{Movie Frames and Optical Flow: Top: frames of the movie. Middle: Optical flow field of the movie. Bottom: Applying optical flow to the frames of the movie. }
	\label{fig:movie}
\end{figure}

\subsection{Histogram matching improves color coherence across scenes}

\begin{figure}
    \begin{center}
        \begin{subfigure}{0.8\linewidth}
            \includegraphics[width=\textwidth]{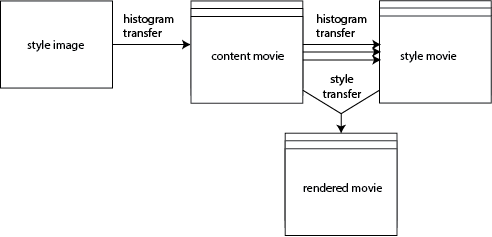}
            \caption{\label{fig:diagram}}
        \end{subfigure}
        \begin{subfigure}{0.8\linewidth}
            \includegraphics[width=\textwidth]{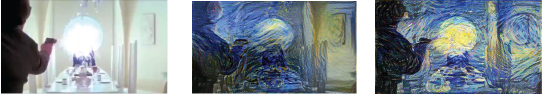}
            \caption{\label{fig:blaster}}
        \end{subfigure}
        \begin{subfigure}{0.8\linewidth}
            \includegraphics[width=\textwidth]{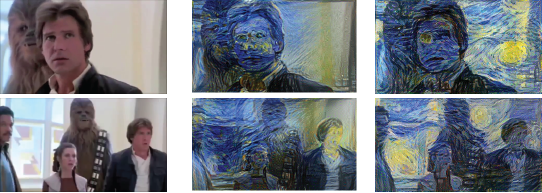}
            \caption{\label{fig:bluehan}}
        \end{subfigure}
    \end{center}
    \caption{\subref{fig:diagram}. In order to allow individual rendered frames to vary in their color palette as they do in the original, we apply histogram matching. First, we transfer the histogram of ``Starry Night'' to the entire movie viewed as an image (i.e., the histogram of the entire movie is matched to that of the painting). Then, we transfer the histogram of each frame of the transformed movie, to a frame of the new ``style movie.'' Thus, each frame of the style movie captures a subset of the painting's original histogram. \subref{fig:blaster}. This allows us to capture brightness changes, as in special effects. From left to right, the original movie frame; the frame rendered using the optical flow approach; and the frame rendered using the optical flow approach with histogram matching. \subref{fig:bluehan}. Similarly, we can preserve the color of objects and characters reappearing over the course of the movie. From left to right, two original frames depicting the same character; the frames rendered using the optical flow approach; and the frames rendered using optical flow and histogram matching.}
    \label{fig:hist_transfer}
\end{figure}

In contrast to a static content image, a movie scene might have colors and textures which change in time. In extending style transfer from images to movies, one might wish to stylistically match the reference image with each frame, or with the movie scene as a whole. For example if the scene is brightly illuminated in one frame and dark in another, one might wish to match the two frames to light and dark colors in the image's palette, respectively, rather than capturing the entire color palette in both. Further, the same object or character might appear as relatively dark or light compared with the background. However, for ease of recognition, one might wish to consistently render them with similar colors.

Matching histograms between movie and image could allow one to capture a large part of the perceptually important color statistics. As an example, we first transfer the RGB histogram of the style image to that of the entire content video. Thus, we allow the colors to vary from frame to frame, while still matching the movie as a whole to the full range of colors. Subsequently, we transfer the histogram of each histogram-transferred video frame back to the original style image in order to allow the colors of the target style to vary in time. This procedure is outlined in Fig. \ref{fig:diagram}. This allows us, for example, to preserve the brightness of special effect blaster fire as in Fig. \ref{fig:blaster}, and the rendering of characters' skin and hair color as in Fig. \ref{fig:bluehan}. The improvement in color coherence comes at the price of a less diverse array of colors and reduction in image quality due to two successive histogram transfers. For related reasons, it also seems to disrupt continuity between frames due to artifactual changes in brightness. This could be improved by using source images with greater color depth. Additionally, similar effects could possibly be obtained without explicit histogram matching using a joint optimization approach as suggested in section \ref{theory}.

\end{document}